\documentclass[letterpaper, 10 pt, conference]{ieeeconf}  

\IEEEoverridecommandlockouts
\IEEEoverridecommandlockouts                              
\usepackage{xcolor}
\usepackage{soul}
\usepackage{cite}
\usepackage{amsmath,amssymb,amsfonts}
\usepackage{graphicx}
\usepackage{textcomp}
\usepackage{xcolor}
\def\BibTeX{{\rm B\kern-.05em{\sc i\kern-.025em b}\kern-.08em
    T\kern-.1667em\lower.7ex\hbox{E}\kern-.125emX}}
\usepackage{subcaption}
\usepackage[colorlinks]{hyperref}
\usepackage{amssymb}
\usepackage{amsmath}
\usepackage[ruled,longend]{algorithm2e}
\usepackage{algpseudocode}
\usepackage{multirow}
\usepackage{booktabs}
\usepackage{caption}

\usepackage{tabularray}
\UseTblrLibrary{booktabs}

\DeclareMathOperator*{\argmax}{arg\,max}

\graphicspath{{figures}}

\begin{document}

\title{Utility AI for Dynamic Task Offloading in the Multi-Edge Infrastructure}

\author{Nazish Tahir \and Ramviyas Parasuraman
\thanks{The authors are with the Heterogeneous Robotics Research Lab (HeRoLab), School of Computing, University of Georgia, Athens, GA 30602, USA. Authors email: {\tt\small \{nazish.tahir,ramviyas\}@uga.edu}}}

\maketitle

\begin{abstract}
To circumvent persistent connectivity to the cloud infrastructure, the current emphasis on computing at network edge devices in the multi-robot domain is a promising enabler for delay-sensitive jobs, yet its adoption is rife with challenges. This paper proposes a novel utility-aware dynamic task offloading strategy based on a multi-edge-robot system that takes into account computation, communication, and task execution load to minimize the overall service time for delay-sensitive applications. Prior to task offloading, continuous device, network, and task profiling are performed, and for each task assigned, an edge with maximum utility is derived using a weighted utility maximization technique, and a system reward assignment for task connectivity or sensitivity is performed. A scheduler is in charge of task assignment, whereas an executor is responsible for task offloading on edge devices. Experimental comparisons of the proposed approach with conventional offloading methods indicate better performance in terms of optimizing resource utilization and minimizing task latency.

\end{abstract}

\section{Introduction}
In robotics, utilizing the cloud for computational offloading is not a novel concept.
Robots may move resource-intensive tasks to cloud nodes with far more resources, which improves job performance, completion time, and battery economy. Offloading enables robot operators to access data from anywhere at any time by providing global storage as well as processing. Furthermore, given the data is maintained in the cloud, robots may be monitored and controlled remotely \cite{tahir2022analog}, facilitating the ability to introduce Cyber-Physical Systems (CPS) that require automated actions in the sense-compute-actuate cycle \cite{spatharakis2022resource}. The dynamic support of CPS is strongly reliant on the timely delivery of the appropriate data placement to the right computing entity \cite{uysal2021semantic}.

\begin{figure}[t]
    \centering
    \begin{center}
    \includegraphics[width=.9\linewidth]{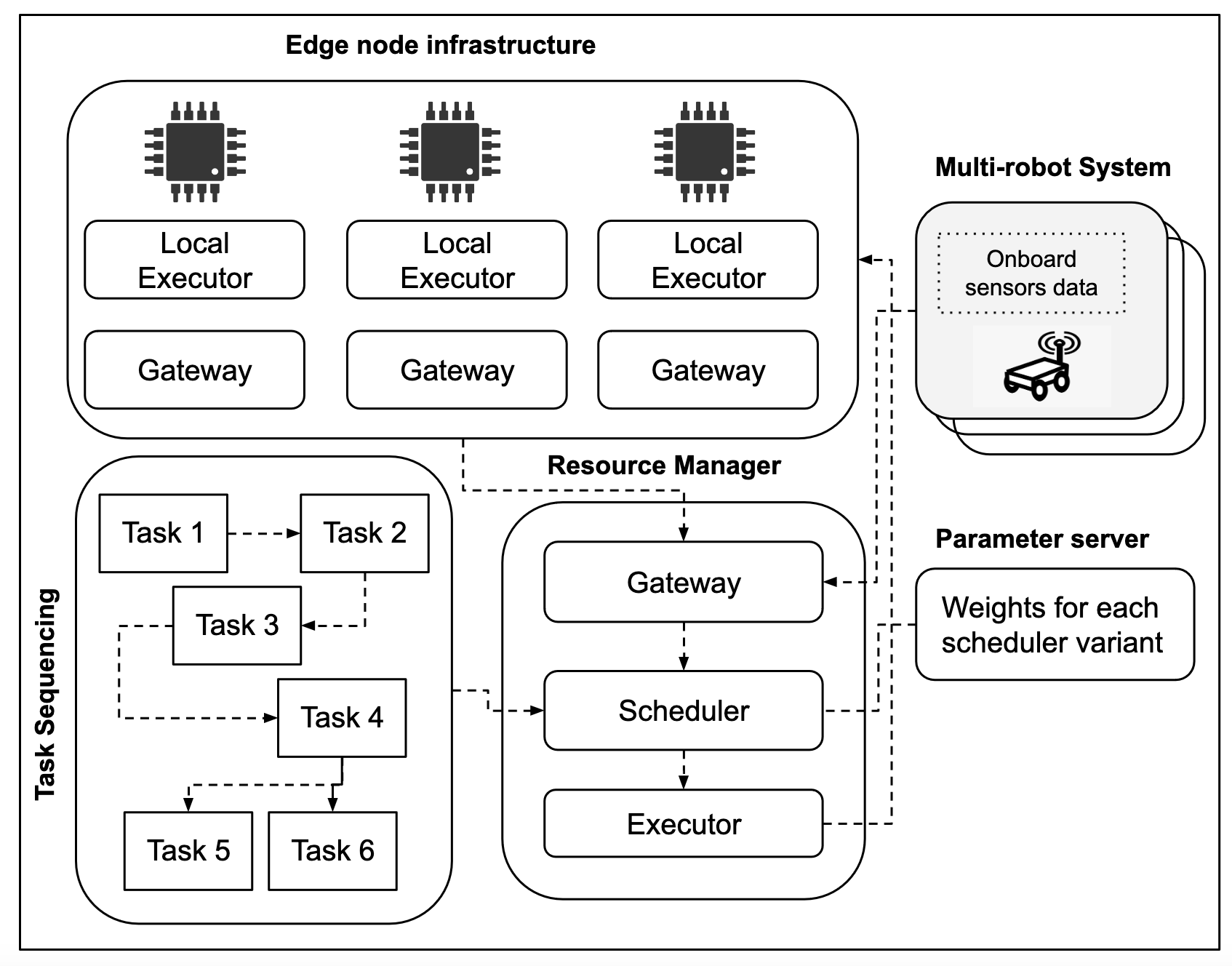}
    \caption{Overview of the proposed utility-aware offloading strategy, illustrating the dynamic task allocation process from multi-robot tasks to multi-edge devices.}
    \label{fig:edge_robotics}
    \vspace{-6mm}
\end{center}
\end{figure}

However, this objective presents its own set of design challenges, including data processing, offloading decision-making, resource allocation, and controller architecture.
Because resource optimization to decrease resource contention while providing performance guarantees is one of the prerequisites for efficient cloud/edge system utilization, an intelligent resource allocation strategy is an imperative need in resource-restricted computational systems. 
The focus of this article is to provide a resource optimization solution with respect to edge robotics applications to make the offloading
decision, which are the topic of extensive study\cite{9289887}, \cite{antevski2018enhancing}, \cite{9677930}. 

Though computational offloading in cloud robotics has made complicated tasks such as map merging, supervised teleoperation, and cooperative navigation possible in a mobile robot scenario, its fundamental disadvantage is the requirement for a persistent connection to an external network \cite{parasuraman2017new}. Cloud computing requires constant access to back-end infrastructure, leading to network congestion and delays, making it inadequate for real-time low-latency and high-quality service applications. \cite{8010408}. 

One promising solution to such a problem is edge computing, whereby edge servers are provisioned at the network edge. Offloading tasks to the edges closer to the robots allows time-sensitive applications to be conducted with shorter latency and lower energy consumption \cite{10.1145/3004010.3004032}. Using network edges enables simultaneous execution of sequential operations, inhibits offloading of larger files, and eliminates data pre-processing, all of which can minimize the overall latency experienced by those applications \cite{9279239}. It also incurs fewer costs than a complete unload to the cloud service. 

The instinctive idea of offloading all tasks to the edge computing system is impractical, leading to instability. As a result, effectively offloading application tasks is necessary yet challenging to balance offloading cost and performance.
Thus, the offloading decisions and resource allocation in order to achieve the trade-off between computational costs and performance efficiency are critical research problems.

This paper proposes a resource maximization offloading for Edge Robotics that aims to reduce offloading costs while ensuring performance guarantees. 
The architectural overview of the proposed dynamic utility-aware multi-robot task offloading strategy is shown in Fig.~\ref{fig:edge_robotics}.
The technique entails dynamically optimizing task offloading on edge servers with the maximum utility to carry out the specific sequence of tasks. Prior to offloading, resources are dynamically assessed based on maximum utility calculations on the real-time network and computing circumstances, as well as job sequencing. 

The proposed approach is evaluated for a multi-robot-multi-edge scenario. It is examined against traditional onboard computation and static offloading techniques to validate the extent to which the optimal resource utilization and task latency minimization objectives have been achieved.
We open source\footnote{\url{https://github.com/herolab-uga/ros-edge-orchestration}} the proposed approach as a ROS package for active use by the research community.

\subsection{Related Work}
Resource allocation with optimization objectives has been widely studied in the literature on robotics applications \cite{dechouniotis2022edge,huang2021efficient}. Since directly offloading computations to cloud-based VMs is impractical owing to network restrictions and latency cost \cite{tahir2023percom}, dividing a demanding workload decreases the computational burden on a single node and assists in battery lifespan. To decrease energy use and computational delay, the authors of \cite{mushunuri2017resource} suggest an optimum partition of the computational load over numerous edge or fog devices.

Similarly, to optimize remote resources in cloud robotics, distributed data flow is used to process complicated data streams generated by the robot's multipurpose sensors, such as map-reduction, using a variety of programming paradigms \cite{4804363}. The authors of \cite{huang2022edge} propose a novel RecSLAM algorithm, a multi-robot laser Simultaneous Localization and Mapping (SLAM) system that focuses on expediting map creation under the robot-edge-cloud architecture. In contrast to traditional Edge-based SLAM \cite{ben2022edge}, this approach provides a hierarchical map fusion technique that decouples the standard SLAM pipeline to spread it across numerous edges via robot data aggregation and edge offloading through graph partitioning. 

The authors of \cite{chinchali2021network} argued that robots should offload only when necessary or highly beneficial and should incorporate network conditions into this calculus. They proposed solutions based on deep reinforcement learning capable of handling diverse network conditions and flexibly trade-off robot and cloud computation. This approach allows robots to intelligently query the cloud for better perception accuracy.

In our previous work \cite{tahir2022analog}, we proposed an analog twin (AT) framework by synchronizing mobility between two mobile robots, where one robot acts as an AT to the other robot. A priority-based supervised bilateral teleoperation strategy is used to offload part of the robotic tasks to the AT without saturating the network conditions.
Similarly, the authors in \cite{xu2020edge} propose a two-layer architecture for visual-based SLAM tracking. They offer a resource-constrained robot with a lightweight version of the visual SLAM, while a considerably more complex and precise version of vSLAM operates on an edge server closer to the robot. The video frames are offloaded to the edge only when necessary thereby decreasing the computational strain on remote resources significantly.

Study in \cite{spatharakis2022resource} optimizes remote resource consumption by presenting a switching set-based control mechanism for robot state estimation that alleviates unnecessary computations and delivers precise navigation to the robot by modeling and quantifying uncertainty. In addition, they tackle the offloading decision-making problem by introducing a utility-based offloading algorithm that considers dynamic network conditions and remote computing resources. Another study \cite{zhang2019resource} investigates a computing efficiency maximization problem in a UAV-enabled MEC system. They jointly optimize the transmission power associated with offloading, the offloading delay, the CPU frequencies, and the trajectory of the UAV to maximize compute efficiency. 
The authors of the \cite{8314091} frame the resource allocation and optimization problem as a Semi-Markov Decision Process (SMDP) to achieve optimum resource allocation in cloud robotics. A reinforcement learning (RL) based autonomous resource allocation scheme is proposed to schedule cloud computing resources efficiently to serve robots and maximize return. 

The authors of \cite{penmetcha2021deep,penmetcha2021predictive} stress the imperative of avoiding under-utilizing a robot's onboard computational resources for efficient offloading. To address this, they formulate the application offloading problem as a Markovian decision process and propose a solution employing deep reinforcement learning through a deep Q-network (DQN) approach. The evaluation conducted on a robot navigation test bed reveals superior performance compared to the baseline long short-term memory (LSTM) algorithm.
In another seminal work \cite{9511225}, authors develop a congestion game-theoretic robotic edge resource allocation mechanism for CPS which minimizes task latency by satisfying the energy constraints of the resources applying to agriculture 4.0 as a use case scenario.

\textbf{Contributions:}
In contrast to the above-mentioned works, the main contributions in the paper are listed below. 
\begin{itemize}
  \item We introduce an innovative offloading strategy for resource optimization, allowing mobile robots to execute resource-intensive tasks beyond onboard capabilities. The approach employs a centralized utility function for dynamic, adaptive offloading decisions based on current remote computational resources and task requirements.
  \item We aim to minimize task latency introduced through subsequent task reassignment between edge devices by introducing a handoff latency minimization approach within the proposed centralized scheduler and executor framework. 
  \item We test the validity of the approach through the following computational-intensive multi-robot tasks: individual robot mapping, navigation, object recognition, classification, and collaborative map merging. 
  \item The proposed CPS is evaluated in terms of map accuracy, navigation task accuracy, object detection's throughput, and the execution time on edge devices. The evaluation includes a comparison of performance metrics under (i) local execution (using robots on-board sensors) (ii) static execution (pre-initialized tasks on edges irrespective of system dynamics changes) (iii) dynamic execution (leveraging proposed utility-aware task offloading mechanism that adapts to current running states and offloads based on task requirements).
\end{itemize}

\section{System Dynamics and Profiling} 
\subsection{Edge Devices and Multirobot System} 
Formally introducing the system architecture, let $\epsilon$ be an edge in edge computing system $\epsilon_{i}$ = $\{\epsilon_{1}, \epsilon_{2}, \epsilon_{3}, ....\epsilon_{m}\}$  where a set of multi-robots $R_{i}$ $\in$ $\{R_{1}, R_{2}, R_{3}, ....R_{n}\}$ can offload their computational tasks $T_{i}$ $\in$ $\{T_{1}, T_{2}, T_{3} ...T_{m}\}$ to the edge servers that are connected via a wireless communication channel in close proximity for faster task execution time.

The multi-robot system is comprised of a group of differential drive robots that communicate with one another using the ROS (Robot Operating System) Communication framework.
To ensure communication between the robots and the edges, ROS services (request/reply communication) or topics (publish/subscribe communication) are employed.
The ROS Master facilitates name registration and lookup within the computation graph, enabling nodes to communicate, exchange messages, and activate services with one another.  
\subsection{Resource Manager}
A local resource manager is an added computational component (a computer, a laptop, or any edge device in the proposed edge robotics architecture) within the local network that performs the computational offloading decision mechanism as well as the task execution algorithm. It also controls ROS communication between edge devices and robots, functioning as a ROS Master. 
\subsection{Edge Computing Resources Profiling}
Three profilers are continually active on the edge devices, monitoring system parameters and delivering them to the local resource manager through ROS topics.  
\subsubsection{Device Profiler} represents the operating states of the edge device and calculates CPU and memory usage percentage.
\subsubsection{Network Profiler} represents the real-time network information, including current throughput calculated by the total MBs transmitted and received calculated from each edge to each individual robot. 
\subsubsection{Task Profiler} represents the requirements of each task as it runs on any remote edge. It tracks the current tasks running on each edge and computes the CPU percentage utilized, the memory size allocated, and network QoS requirements (throughput) for that specific task execution on a particular edge. 

Each of these profilers presents a picture of the changing environment at each time step and records the run-time dynamics of the system as tasks are offloaded or statically pre-initialized to the edge devices.

\section{Dynamic Computational Offloading} 

\subsection{Task Sequencing} 

In the system, we consider a set of tasks $T_{i}$ $\in$ $\{T_{1}, T_{2}, T_{3} ...T_{m}\}$ each of which comes with a predefined sequencing based on task connectivity.

Tasks are sorted by priority of execution in the scheduler to be offloaded to edge devices. Due to interdependence between certain tasks, known as the "condition of order constraint", a successor task cannot be executed before its predecessor task is completed. We call task $T_{i}$ the predecessor of task $T_{j}$, and task $T_{j}$ the successor of task
$T_{i}$. A task is eligible for execution on edge only when its predecessors are already executed; e.g., SLAM is essential to the task of navigation as the map output received from it serves as the input to the ROS navigation stack. The scheduler uses rewards assignments during utility assessment to handle this interdependency. 

In our model, each task is atomic and cannot be further divided. Each computation task can be described in three terms $T_{i}$ $=$ \textlangle $\rho_{cpu_{i}}$, $\rho_{mem_{i}}$, $\rho_{thp_{i}}$ \textrangle. For task $T_{i}$, $\rho_{cpu_{i}}$ is the computing resource demand i.e. CPU capacity required for accomplishing this task quantified by the percentage of CPU and $\rho_{mem_{i}}$ memory usage expressed in percentage, and $\rho_{thp_{i}}$ is the communication resource demand quantified by the required throughput percentage. Note that $\rho_{cpu_{i}}$, $\rho_{mem_{i}}$, $\rho_{thp_{i}}$ are obtained
by the task profiler.

\subsection{Utility-based offloading Strategy}
Our proposed strategy is based on a utility-aware decision-making procedure that quantitatively analyzes and applies the results obtained from the profilers i.e. the current conditions attributed with edge devices obtained, in terms of their associated resource metrics (i.e., availability of networking, storage, and computing resources as well as task requirements at the edge) to a utility function.

The proposed utility function is a convex and non-decreasing(increasing) function of resource usage that reflects benefits or gains derived by the use of the computational/communication entity. By maximizing the sum utility, the offloading scheduler maintains a
balance between competing resource demands of different tasks in a multi-robot system. The utility function calculates the utility score for $\epsilon_{i}$ based on parameters received for $T_{i}$ (by the priority of execution). It calculates \textit{CPU-based utility} for $\epsilon_{i}$ as  
  \begin{equation}
    \begin{aligned}
    UT_{cpu_{\epsilon_{i}}} = \left ( \frac{\Delta_{cpu_{\epsilon_{i}}} - \gamma_{cpu_{\epsilon_{i}}} - \rho_{cpu_{i}} } {\Delta_{cpu_{\epsilon_{i}}}} \right )
    \end{aligned}
   \end{equation}
{where $\Delta_{{cpu}_{\epsilon_{i}}}$ represents the maximum CPU capacity of the edge device $\epsilon_{i}$, $\gamma_{cpu_{\epsilon_{i}}}$ denotes the current CPU usage of $\epsilon_{i}$ and $\rho_{cpu_{i}}$ is the required CPU utilization obtained from task profiler for $T_{i}$.  Similarly, we have \textit{memory-based utility} as follows
   \begin{equation}
    \begin{aligned}
    UT_{mem_{\epsilon_{i}}} = \left ( \frac{\Delta_{mem_{\epsilon_{i}}} - \gamma_{mem_{\epsilon_{i}}} - \rho_{mem_{i}} } {\Delta_{mem_{\epsilon_{i}}}} \right )
    \end{aligned}
   \end{equation}
{where $\Delta_{mem_{\epsilon_{i}}}$ signifies the maximum memory capacity of the edge $\epsilon_{i}$, $\gamma_{mem_{\epsilon_{i}}}$ represents the expected memory utilization which is initialized as the current memory percentage and iterated with the $\rho_{mem_{i}}$, required memory for $T_{i}$. We calculate \textit{network-based utility} by }
   \begin{equation}
    \begin{aligned}
    UT_{net_{\epsilon_{i}}{R_{i}}} = \left ( \frac{\Delta_{thp_{\epsilon_{i}}{R_{i}}} - \gamma_{thp_{\epsilon_{i}}{R_{i}}} - \rho_{thp_{i} } }{\Delta_{thp_{\epsilon_{i}}{R_{i}}}} \right )
    \end{aligned}
   \end{equation}
where $\Delta_{thp_{\epsilon}{R_{i}}}$ signifies the maximum throughput possible between the edge device $\epsilon_{i}$ and robot $R_{i}$, $\gamma_{thp_{\epsilon_{i}}{R_{i}}}$ represents the expected throughput for the task which is initialized as current throughput (transmitted and received from edge $\epsilon_{i}$ to $R_{i}$) and incremented with the $\rho_{thp_{\epsilon_{i}}{R_{i}}}$, the required throughput for $T_{i}$.

}

\subsubsection{System Reward Assignment}
We employ the reward assignment strategy to compute task-based utility.
This reward assignment is dependent on the task sequencing performed previously, as well as the task profiler's knowledge of current tasks executing on the edges. If a predecessor task for the current task $T_{i}$ is already being executed on an edge, it assigns a positive reward to that edge's task-based utility. If its predecessor job is executing on any other edge device, it awards it a negative score (to penalize the network use).

Finally, these task-based utilities are ranked in descending order to obtain each edge's rank as a normalized utility value. A negative penalty may be added to those edges utility that are not currently running the specific task. Once the task-based utility is calculated, it is combined with the other parameters' utility to determine the edge with maximum utility for $T_{i}$

\subsubsection{Weights Variation}
Before calculating the total utility of $\epsilon_{i}$, weights $\omega$ are applied to give importance to the computational or network resources. The weight variants of the proposed scheme entail enforcing maximum weights to the specified parameter, e.g., the CPU-based weight variant assigns maximum weight $\omega_{cpu}$ to the $UT_{cpu_{\epsilon_{i}}}$. Similarly, the memory-based weight variant would assign maximum weight $\omega_{mem}$ to the $UT_{mem_{\epsilon_{i}}}$ as would the network-based weight variant applies maximum weight $\omega_{net}$ to the $UT_{net_{\epsilon_{i} R_{i}}}$. The "all" variant assigns equal weights to all the above-mentioned utilities. Here, all the utilities have been normalized to values between 0 and 1, and the weighing factors have certain restrictions as follows: 
\begin{math}
    \begin{cases}
     0 \leq \omega_{cpu} \leq 1 , \quad 
     0 \leq \omega_{mem} \leq 1, \quad 
     0 \leq \omega_{net} \leq 1 \\
    \omega_{cpu} + \omega_{mem} + \omega_{net} = 1
     \end{cases}
\end{math}

\begin{equation}
    \begin{aligned}
    UT_{total_{\epsilon}} = \omega_{cpu}(UT_{cpu_{\epsilon}}) + \omega_{mem}(UT_{mem_{\epsilon}}) \\ + \omega_{task}(UT_{task_{\epsilon}})
    + \omega_{net}( UT_{net_{\epsilon}{R}})
    \label{eq:2} 
    \end{aligned}
\end{equation}
\begin{equation}
    \begin{aligned}
    \scalebox{1.5}{$\epsilon$}_\mathrm{assigned} = \argmax_\epsilon(UT_\mathrm{total_{\epsilon}{T_{i}}})
    \label{eq:3}
    \end{aligned}
\end{equation}
After the edges utilities $UT_{total_{\epsilon}}$ are calculated with respective weights for $T_{i}$, the utility values are ranked, and the edge $\scalebox{1.5}{$\epsilon$}_\mathrm{assigned}$ with the maximum utility is selected for offloading $T_{i}$.
\subsubsection{Handoff latency avoidance}
The executor uses a handoff latency avoidance technique to compare the task assignment proposed by the scheduler to the existing tasks operating on the edges. If the scheduler proposes re-initializing tasks that overlap with previously existing assignments, the tasks are removed from the allocation list. Thus, the tasks can continue to execute on the same edge device without being reinitialized. Since handoff from one edge device to another requires significant processing power and introduces delay, avoiding the crossover leads to an overall decrease in task latency.
\subsubsection{Completed Task Unloading}
Once the task has been completed and the robot has achieved its objective, the scheduling of its offloading becomes superfluous. The executor eliminates the task upon receiving a prompt from the robot, thereby mitigating computational load on edge servers and minimizing task latency. This technique also helps to optimize the remaining resources, ensuring that the new tasks do not compete for the required resources and that remote resources are not overburdened.

\section{System Implementation} 
Our model consists of a gateway, a scheduler, and an executor, which is in charge of routing the task node to the best available edge device based on the maximum utility value determined for that edge. Fig.~\ref{fig:edge_robotics} presents the thematic diagram showing details of different layers. 

\noindent  \textbf{The Gateway} Before the offloading scheduler starts to work, the gateway is initiated at the local resource manager. The device and network profilers on each edge device should start to profile and publish the data to related topics so that the resource manager can process the data. The Gateway keeps track of real-time data from each edge device, including CPU usage, network utilization, memory usage, and the current tasks running in a centralized manner, and performs preprocessing to keep the most updated data about the current state of the system. This information is constantly published on a ROS topic so that distributed information on the current state of the system is always available to assist in scheduling and executing offloading mechanisms.

\noindent \textbf{The Scheduler}
The scheduler, also located on the Resource Manager, performs the task sequencing either through task connectivity or task sensitivity approach, iterates through the sequence of the tasks, analyzes the current stream of data received from the Gateway, updates the current state by evaluating the required state and calculates the utility value of each edge device based on the utility-based offloading strategy described in detail in section IV. The scheduler is responsible for calculating utility values for each edge device, system reward assignment, and weight variation. After performing these tasks, the scheduler churns out the edge with the maximum utility value for each sequenced task and proposes the task assignment to respective edges to the executor. The proposed algorithm performed at the scheduler is presented in Alg.~\ref{alg:scheduler}

\begin{algorithm}[t]
  \caption{Scheduler}
  \label{alg:scheduler}
  \nl Perform the task sequence $seq_{T_{i}}$\\
  \nl 
    \ForEach{$\epsilon \in \{1, 2, \dots, N\}$}{
      get $\gamma_{cpu_{\epsilon_{i}}}$, $\gamma_{mem_{\epsilon_{i}}}$, $\gamma_{thp_{\epsilon_{i}}{R_{i}}}$ and $T_{curr_{\epsilon_{i}}}$\\
      \Call {Edge\_Assignment()}\\
    }
    \nl \textbf{Function} {Edge\_Assignment()}\\
        \nl  initialize metrics\\
        \nl  \ForEach{$T_{i} \in \{T_{1}, T_{2},\dots,T_{N}\}$}{
            Get $\rho_{cpu_{i}}$, $\rho_{mem_{i}}$, $\rho_{thp_{i}}$\\
            $\scalebox{1.5}{$\epsilon$}_\mathrm{assigned} $ = \Call {Calculate\_Utility($T_{i}$, $\rho_{i}$)}\\
            Publish ($T_{i}$ : $\scalebox{1.5}{$\epsilon$}_\mathrm{assigned}$ ) to the Executor\\
    }

    \nl \textbf{Function} {Calculate\_Utility($T_{i}$, $\rho_{i}$)}\\
    \nl set $\Delta_{cpu_{\epsilon}}$, $\Delta_{mem_{\epsilon}}$, $\Delta_{thp_{\epsilon}{R}}$\\
    \nl $UT_{cpu}$ = $\Delta_{cpu_{\epsilon}}$ - $\gamma_{cpu_{\epsilon}}$ - $r\rho_{cpu_{\epsilon}}$ / $\Delta_{cpu_{\epsilon}}$ \\
    \nl $UT_{mem}$ = $\Delta_{mem_{\epsilon}}$ - $\gamma_{mem_{\epsilon}}$ - $\rho_{mem_{\epsilon}}$ / $\Delta_{mem_{\epsilon}}$ \\
    \nl $UT_{net}$ = $\Delta_{thp_{\epsilon}{R}}$ - $\gamma_{thp_{\epsilon}{R}}$ - $\rho_{thp_{T_{i}}}$ / $\Delta_{thp_{\epsilon}{R}}$ \\
    \nl $UT_{tasks}$ =  Task\_Utility($T_{i}$) \\
    \nl set $\omega_{param}$ \\
    \nl $UT_{total_{\epsilon}} = \omega_{cpu}(UT_{cpu_{\epsilon}}) + \omega_{mem}(UT_{mem_{\epsilon}})  + \omega_{net}( UT_{net_{\epsilon}{R}}) + \omega_{task}(UT_{task_{\epsilon}})$ \\
    \nl List ${T_{i}:\epsilon}$ = $UT_{total_{\epsilon}}$ \\
    \Return ${T_{i}:\epsilon}$ = max($UT_{total_{\epsilon}}$) \\
    \nl \textbf{Function} {Task\_Utility($T_{i}$)}\\
    \nl get $seq_{T_{i}}$, $T_{curr_{\epsilon}}$ \\
    \nl \uIf{$seq_{T_{i}}$ $\in$ $T_{curr_{\epsilon}}$}{
        $UT_{tasks}$ = +1\\
    }
    \Else{
    $UT_{tasks}$ = -1 \\
    }
    \Return $UT_{tasks}$

\end{algorithm}
\noindent \textbf{The Executor} 
The edge assigned to each task is subsequently transferred to the executor running on the local resource manager. The executor maintains a final allocation layer where it checks a list of task nodes previously assigned, compares it to the recent assignment received by the scheduler, and updates the list to minimize overlapped offloading as proposed in the handoff latency avoidance mechanism. The executor is responsible for maintaining connections with the edge devices, and after establishing a good connection, it offloads the task on the respective edge. Once the task for any of the robots in the multi-robot system is finished, the executor removes it from the edge to conserve computing and storage resources and avoid rescheduling tasks no longer needed to the edges through the completed tasks unloading mechanism. The proposed algorithm performed at the executor is presented in Alg.~\ref{alg:executor}

\begin{figure}[t]
    \centering
    \includegraphics[width=.24\textwidth]{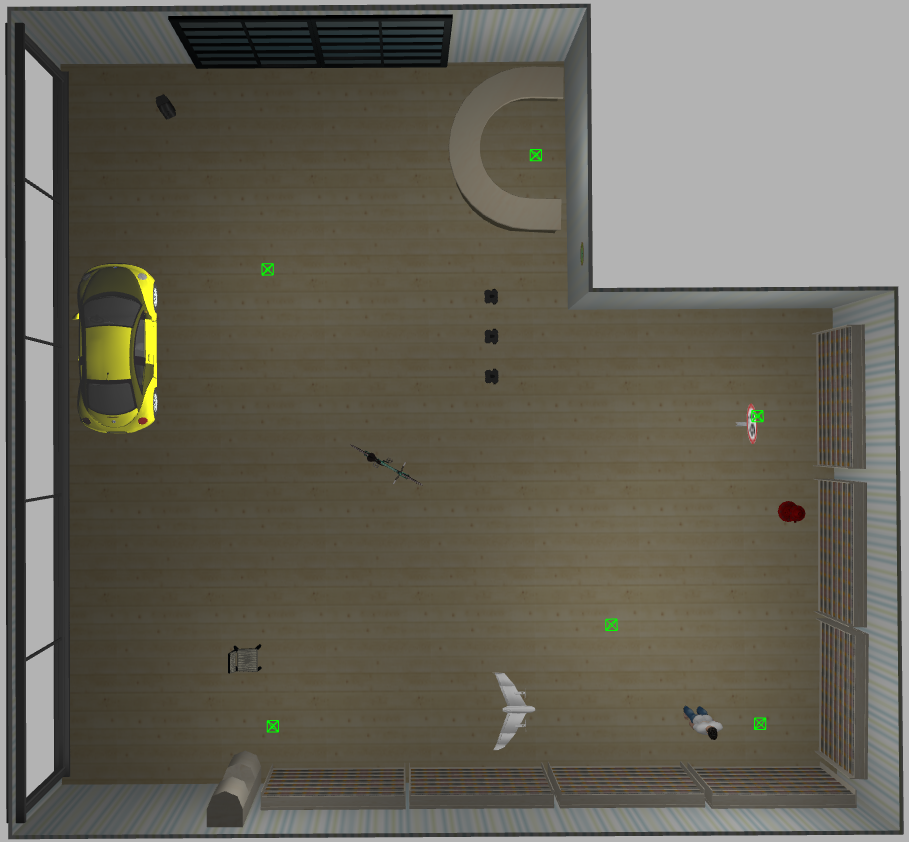}
    \includegraphics[width=.22\textwidth]{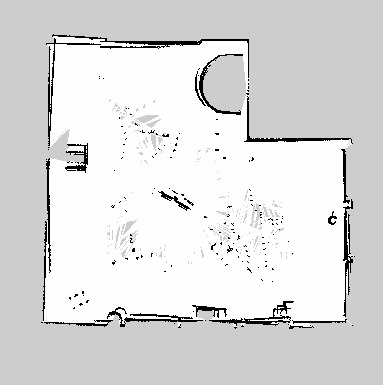}
    \caption{Experiment setup of the simulations in ROS Gazebo (left): Gazebo world with random distinct multiple objects placed to perform object detection. (right): Shows the SLAM mapping obtained by multiple robots navigation upon completing the task collaboratively.}
    \label{fig:simulation-map}  
    \vspace{-2mm}
\end{figure} 

\begin{algorithm}[t]
  \caption{Executor}
  \label{alg:executor}
  \nl Obtain ($Ti$ : $\scalebox{1.5}{$\epsilon$}_\mathrm{assigned}$) from Scheduler\\
  \nl Populate a recent assignment list $List$ = \{i:$\scalebox{1.5}{$\epsilon$}_\mathrm{assigned}$\}  \\
  \nl Compare with previous assignment  list and remove overlap \\ 
  \nl Perform task status check \\
  \nl Obtain command for $T_{i}$ \\
  \Call{} {Task\_Offload($T_{i}$,$\scalebox{1.5}{$\epsilon$}_\mathrm{assigned}$} \\
   \nl \textbf{Function} {Task\_Offload($T_{i}$, $\scalebox{1.5}{$\epsilon$}_\mathrm{assigned}$)}{}\\
    \nl  login (hostname, username, password) \\
    \nl sendline($T_{i}$ command) \\
    \nl logout \\ 
    \nl \textbf{End function} \\
\end{algorithm}

\section {Experiment Setup}
Using the ROS-based robotics simulator Gazebo, we created a multi-robot simulation environment using three Turtlebot3 robots and three edge devices. The system employed for simulation is equipped with Intel® Core™ i7-8565U CPU @ 1.80GHz × 8 and 16-GB memory with Ubuntu Server with ROS Noetic. The multi-robot system is integrated via a wireless network with three edge devices running ARMv8 Processor rev 1(v8l) x 4 architectures, with 4GB memory and NVIDIA Tegra X1 nvgpu integrated each with Ubuntu 20.04.4 LTS operating system running ROS Noetic.

The experimental scenario utilizes mapping performed through Gmapping SLAM, navigation by using ROS navigation, object detection, and classification through using YOLO v5 and Kinect depth cameras, and map merging by ROS package multirobot\_map\_merge. 
For the evaluation of our proposed scheduler, we implemented multi-robot cooperative object detection and tracking, where individual robots are tasked with detecting and collecting objects in their environment and localizing the objects. The task is completed once all the robots have collaboratively detected and localized all distinct objects in their proximate space. Until then, robots keep exploring the space, traversing the waypoints given using ROS navigation stack.  

\noindent \textbf{Task Implementation:} The tasks selected for execution on the multi-robot system through edge-based remote computational infrastructure are 
\textit{(i)} \textit{ Individual Robot mapping} \textit{(ii)} \textit{ Navigation } \textit{(iii)} \textit{ Object Detection and Classification} \textit{(iv)} \textit{ Map merging}. The task is tested in two scenarios: 
\begin{itemize}
\item Scenario 1: involves robots performing individual mapping through SLAM and merging the map collaboratively through onboard processors while offloading only object detection and mapping through the proposed utility-aware scheduler. 
\item Scenario 2: involves individual mapping and map-merging statically initialized at the edge devices while the scheduler takes care of scheduling object detection and navigation to the already constrained edge devices. 
\end{itemize}

\noindent \textbf{Comparison schemes:} We compare the performance of three computational offloading schemes in terms of task latency and optimal use of remote resources. 
\begin{itemize}
    \item \textit{Local execution} where the robot never invokes the resource manager for offloading decisions and relies on the onboard processors for task execution. 
    \item \textit{Static execution} where the tasks are offloaded to the edges statically, and no handoff or change occurs in their assignment until the tasks are fully executed.
    \item \textit{Dynamic execution} where the tasks are offloaded to the edges according to the proposed utility-based computational offloading strategy. We test the following weight variants of the dynamic offloading scheme: CPU-based, memory-based, network-based, and all variants. 
\end{itemize}

\noindent \textbf{Evaluation metrics:} To evaluate the performance of different offloading schemes, we define the following metrics
\begin{itemize}
\item CPU utilization: It is measured as the peak CPU usage rate of a computational resource. 
\item RAM Memory utilization: It is measured as the total memory usage of a computational resource. 
\item Network Throughput (TP): It is measured as the number of megabytes transmitted and received per second at a computing resource. 
\item Task Latency (TL): It is measured as the total execution time of the given task. 
\item SSIM: The Structural Similarity Index Measurement (SSIM) is used for assessing the accuracy of the mapping performed by comparing it with the ground truth map referred to in Fig.~\ref{fig:simulation-map}.
\item Frames Per Second (FPS): Upon scheduling the object detection task, the processing of the frames is affected according to the edges' computational capability. We record this FPS for the stream of images coming from the robot's camera frame for object detection.
\end{itemize}

\begin{table*}[t]
    \centering
    \caption{Performance results of evaluation metrics in Scenarios 1 and 2. Bold indicates the best performance.}
    \label{table:performance_sc1}
    \resizebox{\linewidth}{!}{%
    \begin{tabular}{|r|l|l|l|l|l|l|l|l|l|l|l|l|l|l|l|l|}
    \hline
               \multirow{3}{*}{\textbf{Parameter}} & \multirow{3}{*}{\textbf{Local}}  & \multicolumn{3}{|c|}{\textbf{Static }} & \multicolumn{12}{|c|}{\textbf{Dynamic}}
               \\ \cline{3-17}

        & ~ & ~ & ~ & ~ & \multicolumn{3}{|c|}{\textbf{CPU Variant}} & \multicolumn{3}{|c|}{\textbf{Network Variant}} & \multicolumn{3}{|c|}{\textbf{Memory Variant}} & \multicolumn{3}{|c|}{\textbf{All Variant}} \\ \cline{3-17}
        
       & ~ & Edge 1 & Edge 2 & Edge 3 & Edge 1 & Edge 2 & Edge 3 & Edge 1 & Edge 2 & Edge 3 & Edge 1 & Edge 2 & Edge 3 & Edge 1 & Edge 2 & Edge 3 \\ \hline

      \multicolumn{17}{|c|}{\textbf{\Large Scenario 1}}  \\ \hline

        \textbf{CPU (\%) Mean} & 99.80 & 99.95 & 99.87 & 99.90 & 85.06 & 81.94 & 99.72 & 81.47 & 80.49 & 99.87 & \textbf{77.63} & \textbf{55.20} & \textbf{99.82} & 98.81 & 93.52 & 97.57 \\  
        \textbf{  STD} & 0.06 &  0.03 & 0.13 & 0.04 & 8.41 & 12.46 & 0.29 & 11.54 & 14.03 & 0.09 & \textbf{14.72} & \textbf{20.39} & \textbf{0.19} & 1.84 & 9.23 & 4.35 \\  \hline

        \textbf{RAM (\%) Mean} & 80.29 & 64.54 & 61.00 & 57.90 & 51.88 & 50.31 & 62.76 & \textbf{47.75} & \textbf{58.44} & \textbf{72.25} & 43.54 & 56.33 & 84.10 & 51.65 & 67.84 & 68.95  \\  
        \textbf{STD} & 3.99 & 15.44 & 14.11 & 20.24 & 7.09 & 8.39 & 7.57 & \textbf{6.14} & \textbf{13.93} & \textbf{6.54} & 7.21 & 5.43 & 7.91 & 3.67 & 6.83 & 10.72 \\  \hline

        \textbf{Throughput (MBps) Mean} & 1.55 & 64. 94 & 64.15 & 62.87 & 51.10 & 52.45 & 55.80 & 56.45 & 58.04 & 64.60 & \textbf{ 62.75} & \textbf{32.54} & \textbf{60.39} & 63.48 & 57.58 & 44.58 \\ 
        
        \textbf{STD} & 0.83 & 4.57 & 6.31 & 5.03 & 14.07 & 7.61 & 11.84 & 7.03 & 16.35 & 16.34 & \textbf{6.40} & \textbf{11.85} & \textbf{15.68} & 17.22 & 12.96 & 17.00 \\ \hline
 
        \textbf{YoLov5 FPS (Hz) Mean}  & 8.62 & 3.57 & 3.68 & 3.53 & 2.71 & 2.13 & 2.84 & 3.30 & 2.56 & 3.26 & 2.91 & 2.37 & 2.64 & \textbf{3.30} & \textbf{3.03} & \textbf{3.16} \\ 
        \textbf{STD } & 4.28 & 0.49 & 0.22 & 0.29 & 0.55 & 0.55 & 0.65 & 1.47 & 1.77 & 1.19 & 0.21 & 0.46 & 0.44 & \textbf{0.50} & \textbf{0.86} & \textbf{0.80} \\  \hline
        
        \textbf{SSIM (\%) Mean} & 0.73  &	\multicolumn{3}{|c|}{0.73} &	\multicolumn{3}{|c|}{0.73}  &	\multicolumn{3}{|c|}{0.74}	& \multicolumn{3}{|c|}{0.74} &	\multicolumn{3}{|c|}{\textbf{0.75}}\\ 
        \textbf{STD} &  \textbf{0.02} &	\multicolumn{3}{|c|}{0.01} &	 \multicolumn{3}{|c|}{0.01}&	\multicolumn{3}{|c|}{0.02}	& \multicolumn{3}{|c|}{0.01}&	\multicolumn{3}{|c|}{\textbf{0.02}}\\ \hline

        \textbf{Task Latency (mins) Mean} &14.40 & \multicolumn{3}{|c|}{9.93} & \multicolumn{3}{|c|}{8.88} &	\multicolumn{3}{|c|}{8.84}  &	\multicolumn{3}{|c|}{9.72} 	 & \multicolumn{3}{|c|}{\textbf{6.93}}\\ 
        \textbf{STD} &7.78 & \multicolumn{3}{|c|}{2.15}	 & 	\multicolumn{3}{|c|}{2.45} &	\multicolumn{3}{|c|}{3.45}  &	\multicolumn{3}{|c|}{1.69}	 &  \multicolumn{3}{|c|}{\textbf{3.42}} \\ \hline
     \multicolumn{17}{|c|}{\textbf{\Large Scenario 2}}  \\
    \hline

        
        
        \textbf{CPU (\%) Mean} & 99.8 & 99.5  & 99.73  & 99.5 & 87.7  & 83.5  & 99.7  & \textbf{87.6} &  \textbf{78.13} & \textbf{99.94} & 85.12  & 82.48  & 99.93  & 88.34  & 82.5  & 99.8  \\  
        \textbf{  STD} & 0.05 &  0.87 &  0.31 & 0.45 &  11.63 &  14.7 & 0.26 & \textbf{10.26} &  \textbf{15.74} &  \textbf{0.02} &  9.37 &  25.82 &  0.03 &  12.25 &  10.5 &  0.04 \\  \hline

        \textbf{RAM (\%) Mean} & 80.29 & 65.95  & 63.2 & 70.4 & 59.5  & 56.4  & 66.63 & \textbf{56.1} & \textbf{50.8} & \textbf{66.23}  & 53.77  & 53.34  & 66.5  & 57.23  & 53.54 & 63.38 \\  
        \textbf{STD} &  &  0.3 &  0.5 &  16.4 &  7.07 & 7.07 &  4.98 &  \textbf{4.5} & \textbf{ 5.66} &  \textbf{8.08} &  5.24 &  9.90 &  11.04 &  6.98 &  5.59 &  2.86 \\  \hline

        \textbf{Throughput (MBps) Mean} & 0.25  & 74.94  &	70.26 	&	73.60 &	\textbf{79.06}	& \textbf{70.05}	&	\textbf{82.37} &	58.56	&		59.12		& 	82.90	&	44.87 	&	37.51  &	69.42	&	51.47 &	46.19	&	68.06  \\ 
        \textbf{STD} & 0.22 &  15.13 &	14.70	&	 16.11	&	 \textbf{21.75}	& \textbf{14.74}	&	 \textbf{18.21}  &	 22.22	&		22.59		& 	33.60	&	14.86	&	 27.46  &		26.93	&	13.56	&	 10.18	&		28.45 \\ \hline

        \textbf{YoLov5 FPS (Hz) Mean} & 8.62 
        &         3.04&          3.04 &	         2.50&         \textbf{2.63}  &         \textbf{3.96} &	      \textbf{3.41} &        2.47  &        2.89 &	        3.29&	        3.01 &	        2.45&	        2.18  &        1.78  &        1.40  &         2.53 \\ 
        \textbf{STD } & 4.28  
        &         1.27 &          1.18 &	         0.98 &         \textbf{0.93} &        \textbf{1.15} &	      \textbf{0.55} &       1.05 &       1.73 &	       1.78 &	      0.69 &	       1.00&	       0.96 &        0.45 &       0.26 &         0.92   \\  \hline
        
        \textbf{SSIM (\%) Mean} & 0.73  &	\multicolumn{3}{|c|}{0.74} &	\multicolumn{3}{|c|}{0.74}  &	\multicolumn{3}{|c|}{0.72}	& \multicolumn{3}{|c|}{\textbf{0.75}} &	\multicolumn{3}{|c|}{0.74}\\ 
        \textbf{STD} &  0.02 &	\multicolumn{3}{|c|}{0.02} &	 \multicolumn{3}{|c|}{0.01}&	\multicolumn{3}{|c|}{0.01}	& \multicolumn{3}{|c|}{\textbf{0.03}}&	\multicolumn{3}{|c|}{0.03}\\ \hline

        \textbf{Task Latency (mins) Mean} &14.42 & \multicolumn{3}{|c|}{10.08} & \multicolumn{3}{|c|}{9.85} &	\multicolumn{3}{|c|}{9.88}  &	\multicolumn{3}{|c|}{\textbf{6.21}} 	 & \multicolumn{3}{|c|}{6.84}\\ 
        \textbf{STD} &7.78 & \multicolumn{3}{|c|}{1.98}	 & 	\multicolumn{3}{|c|}{1.78} &	\multicolumn{3}{|c|}{0.36}  &	\multicolumn{3}{|c|}{\textbf{2.92}}	 &  \multicolumn{3}{|c|}{2.93} \\ \hline

    \end{tabular}
    }
\end{table*}

\begin{figure*}[ht]

\captionsetup{justification=centering,margin=2cm}
\centering
    \begin{subfigure}[t]{0.32\textwidth}
    \centering
    \includegraphics[width=\textwidth]{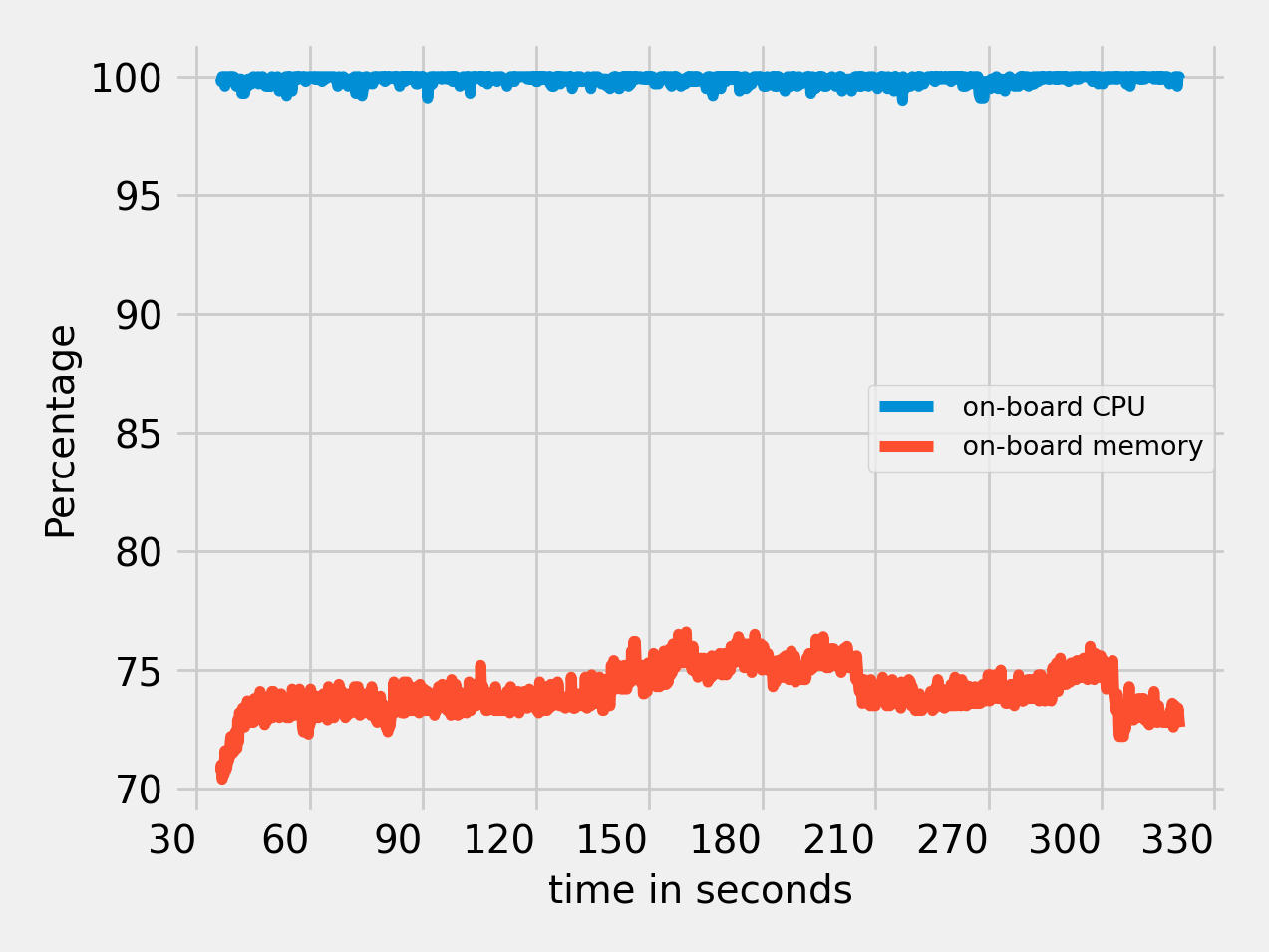}
    \caption{Local Computation } \label{ts_local_sc1}
\end{subfigure}
\begin{subfigure}[t]{0.32\textwidth}
    \centering
    \includegraphics[width=\textwidth]{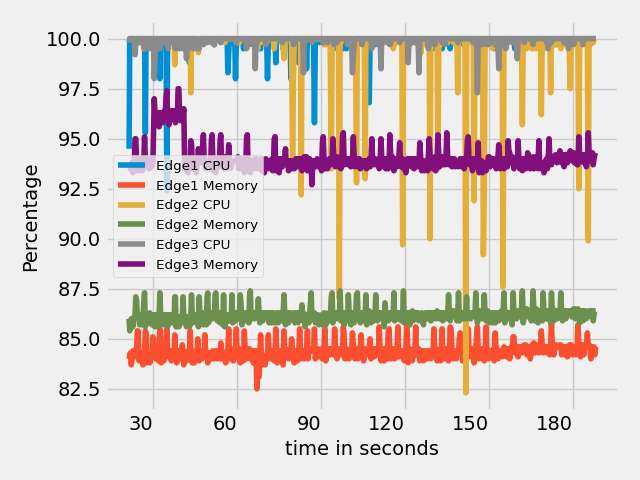}
    \caption{Static Offloading} \label{ts_static_sc1}
\end{subfigure} 
 \begin{subfigure}[t]{0.32\textwidth}
    \centering
    \includegraphics[width=\textwidth]{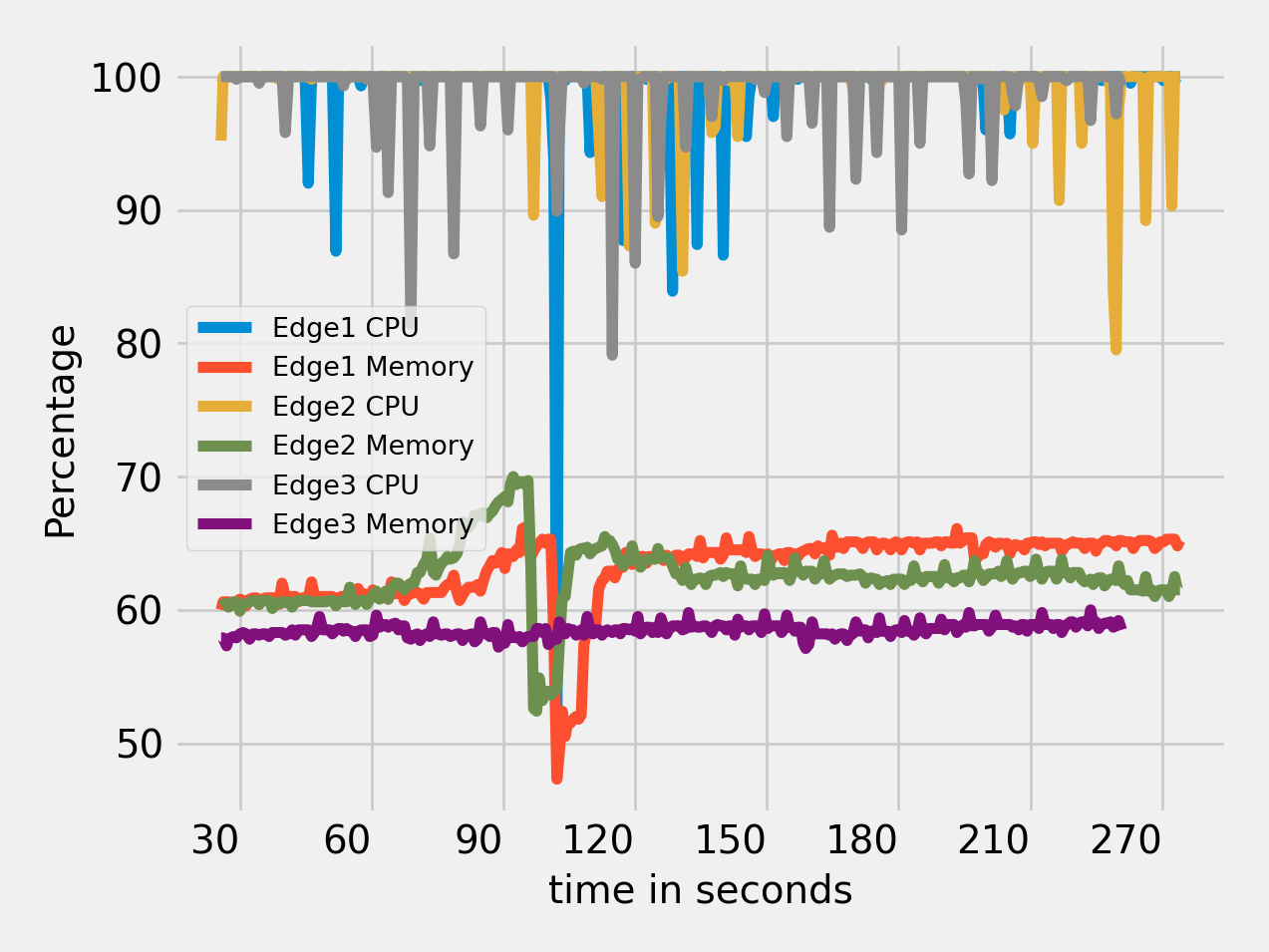}
    \caption{Dynamic Offloading (CPU-variant)} \label{ts_dy_cpu_sc1}
\end{subfigure}
\begin{subfigure}[t]{0.32\textwidth}
    \centering
    \includegraphics[width=\textwidth]{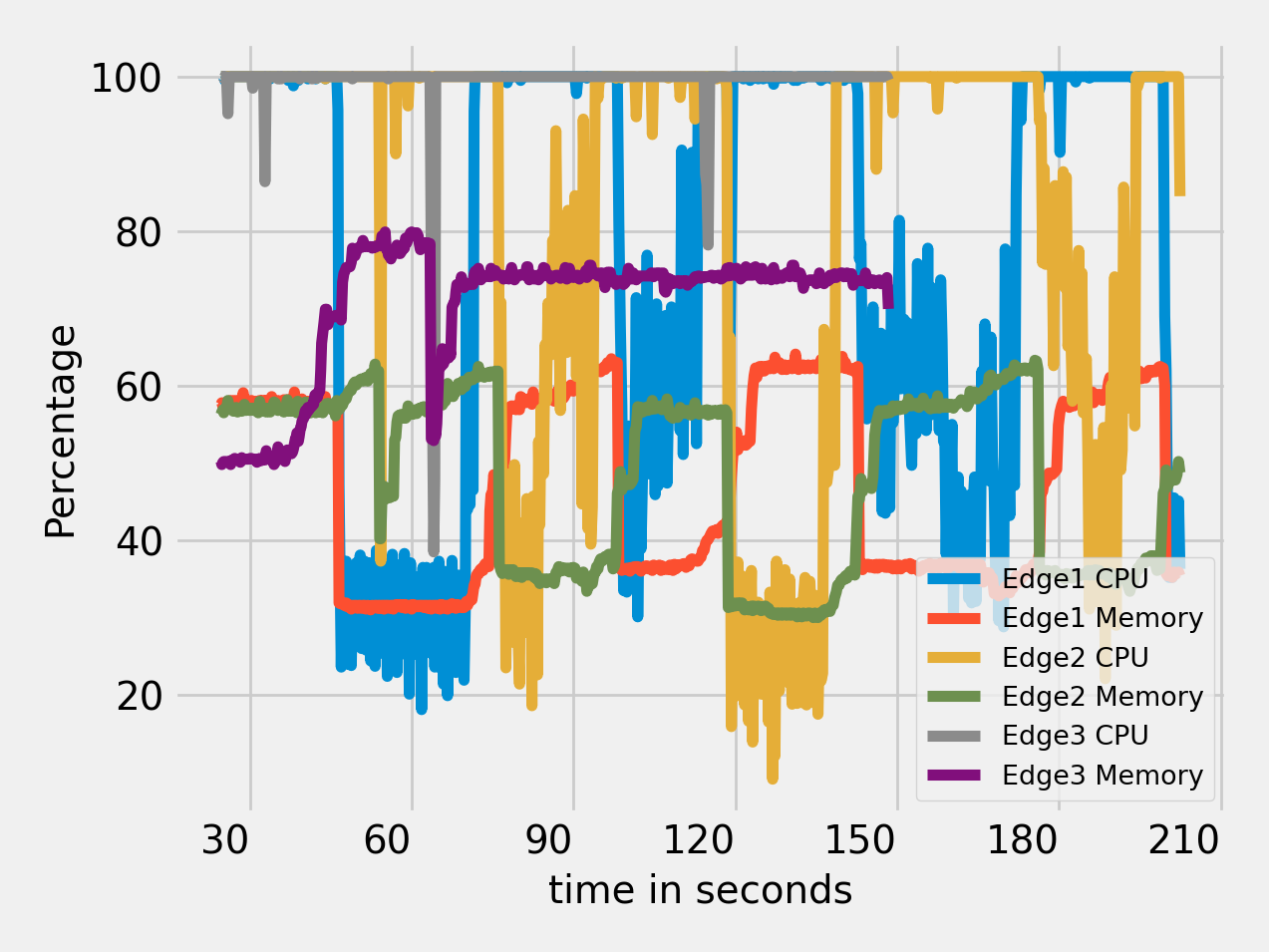}
    \caption{Dynamic Offloading (Memory-variant)} \label{fig:ts_dy_mem_sc1}
\end{subfigure}
\begin{subfigure}[t]{0.32\textwidth}
    \centering
    \includegraphics[width=\textwidth]{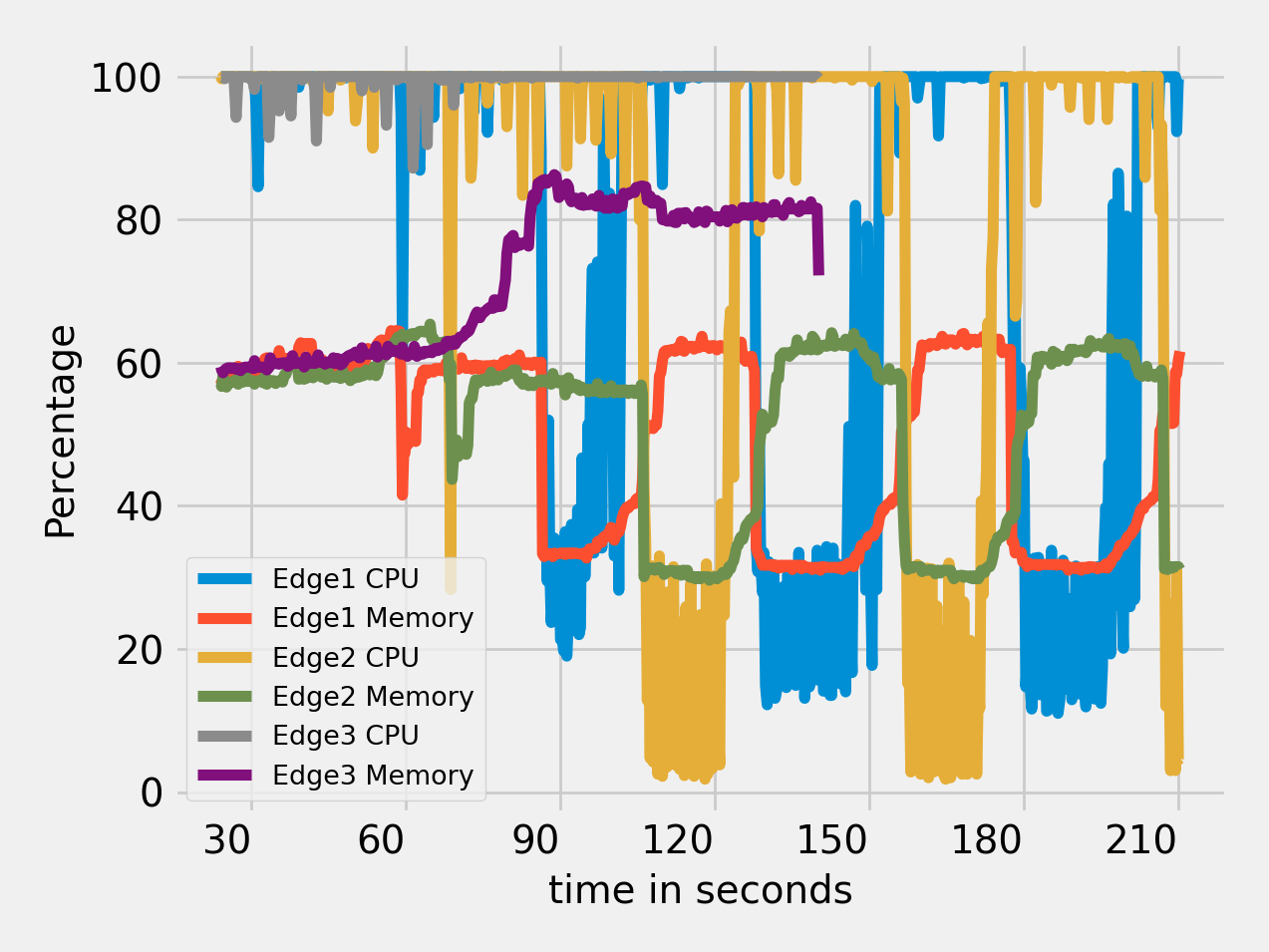}
    \caption{Dynamic Offloading (Network-variant)} \label{ts_dy_net_sc1}
\end{subfigure}
\begin{subfigure}[t]{0.32\textwidth}
    \centering
    \includegraphics[width=\textwidth]{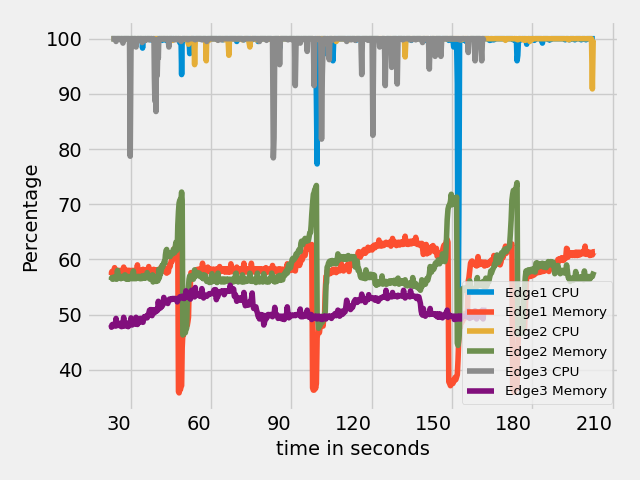}
    \caption{Dynamic Offloading (All-variant)} \label{ts_dy_all_sc1}
\end{subfigure}
\caption{Scenario 1: Resource changes of CPU and memory across devices in a sample trial.}
\label{fig:timeseries_sc1}
\vspace{-2mm}
\end{figure*}

\begin{figure*}[ht]
\captionsetup{justification=centering,margin=2cm}
\centering
    \begin{subfigure}[t]{0.31\textwidth}
    \centering
    \includegraphics[width=\textwidth]{local_sc1.png}
    \caption{Local Computation } \label{ts_local_sc2}
\end{subfigure} 
\begin{subfigure}[t]{0.31\textwidth}
    \centering
    \includegraphics[width=\textwidth]{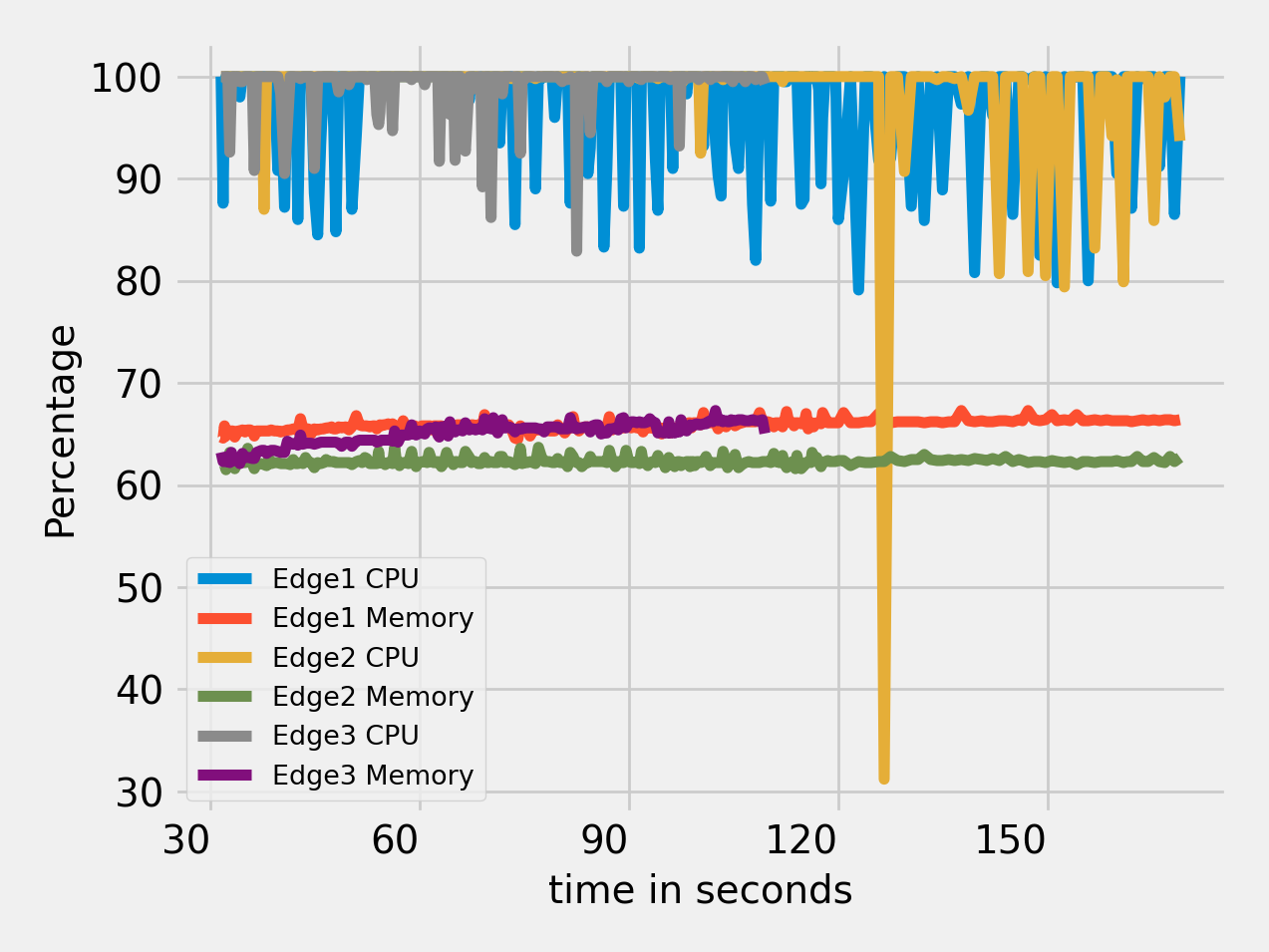}
    \caption{Static Offloading} \label{ts_static_sc2}
\end{subfigure} 
 \begin{subfigure}[t]{0.31\textwidth}
    \centering
    \includegraphics[width=\textwidth]{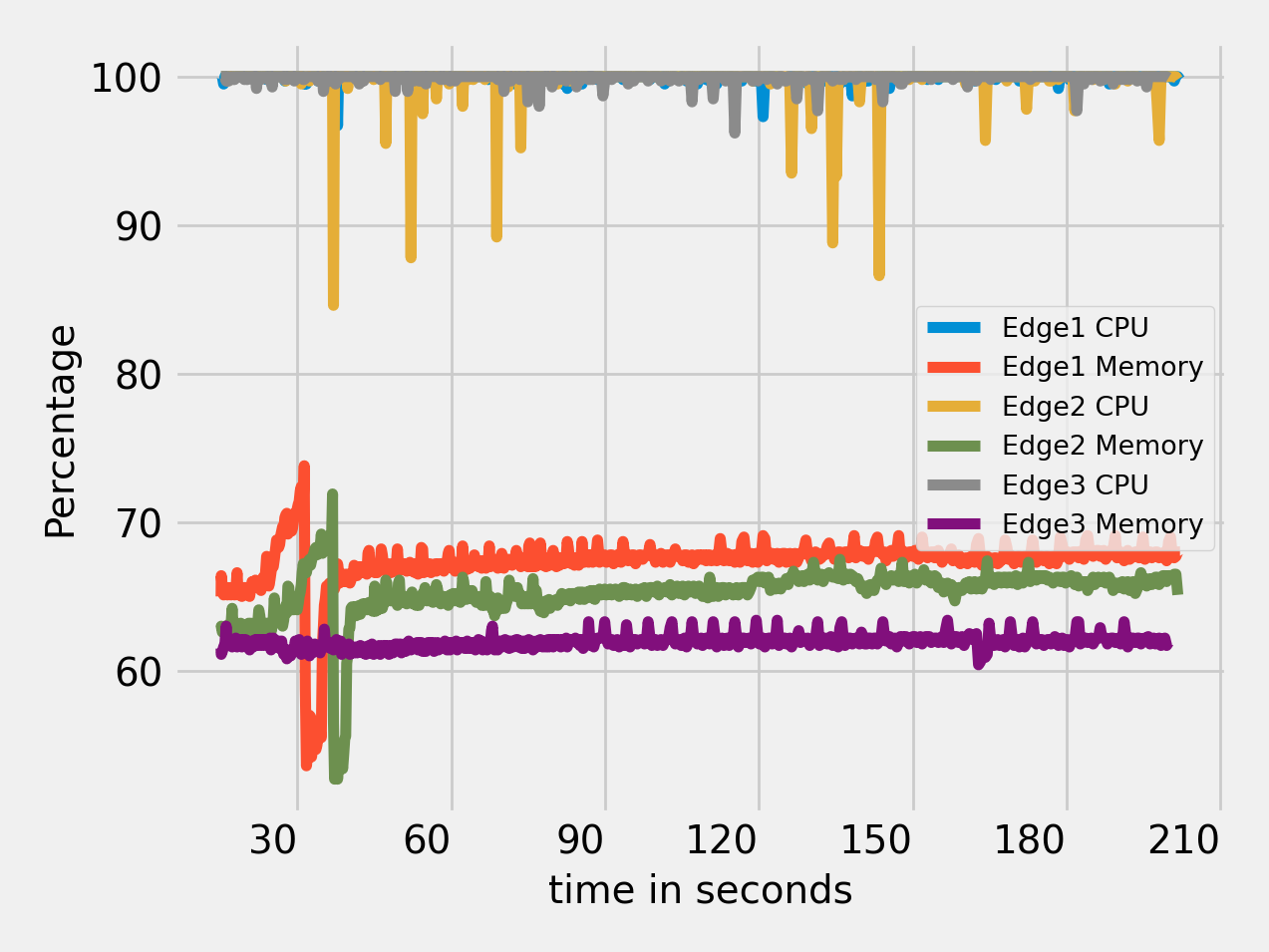}
    \caption{Dynamic Offloading (CPU-variant)} \label{ts_dy_cpu_sc2}
\end{subfigure}
\begin{subfigure}[t]{0.31\textwidth}
    \centering
    \includegraphics[width=\textwidth]{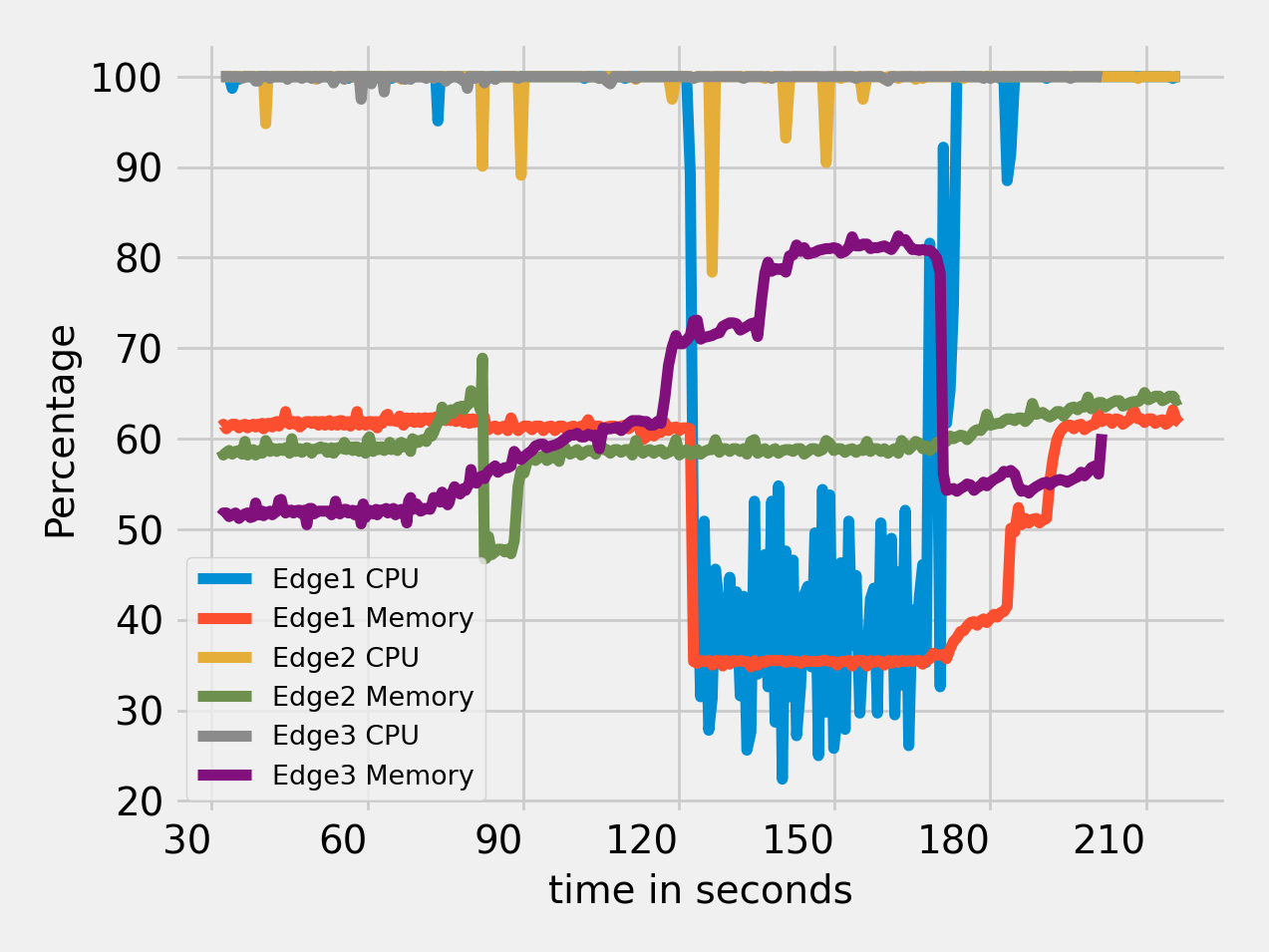}
    \caption{Dynamic Offloading (Memory-variant)} \label{ts_dy_mem_sc2}
\end{subfigure}
\begin{subfigure}[t]{0.31\textwidth}
    \centering
    \includegraphics[width=\textwidth]{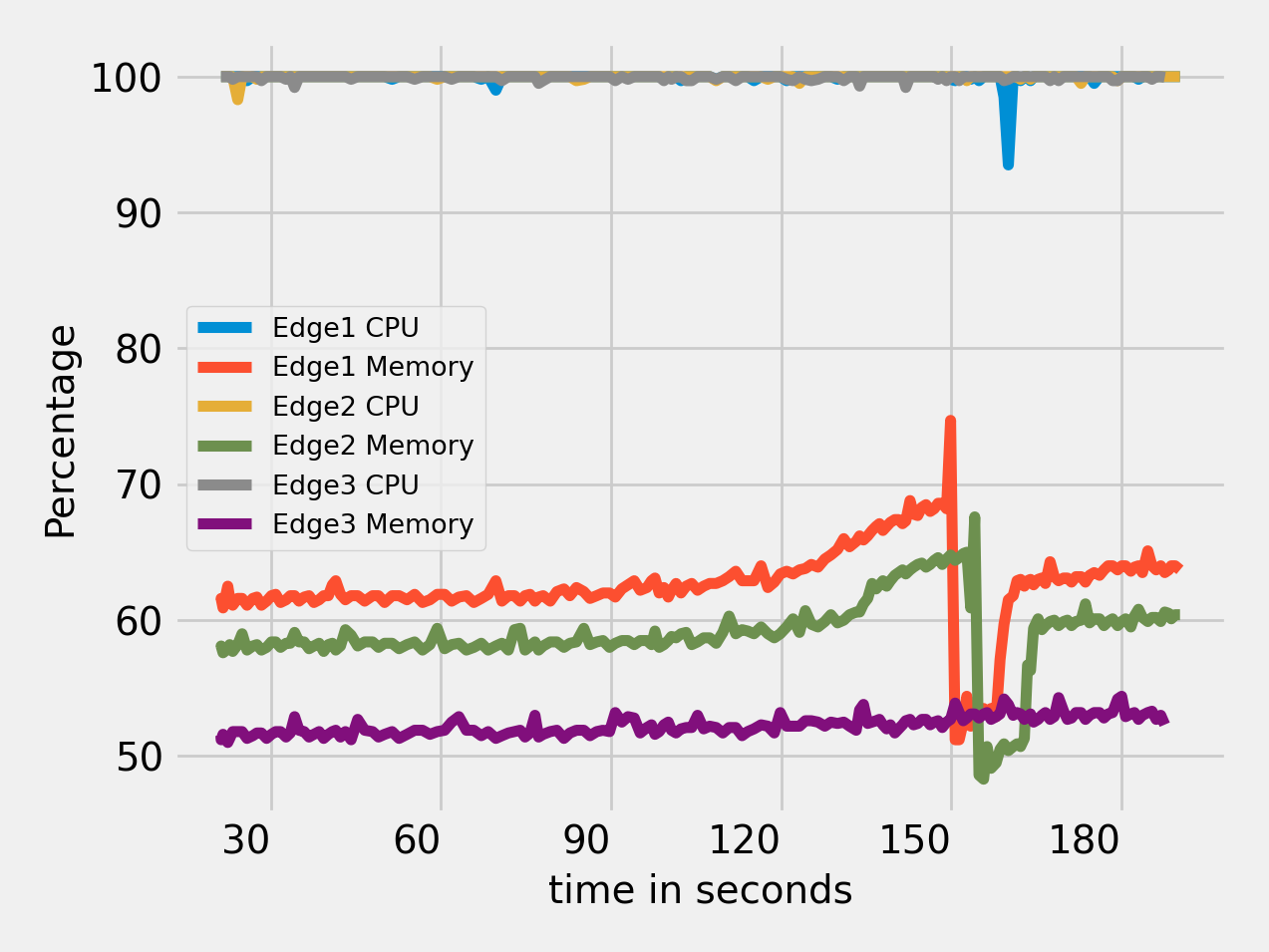}
    \caption{Dynamic Offloading (Network-variant)} \label{ts_dy_net_sc2}
\end{subfigure}
\begin{subfigure}[t]{0.31\textwidth}
    \centering
    \includegraphics[width=\textwidth]{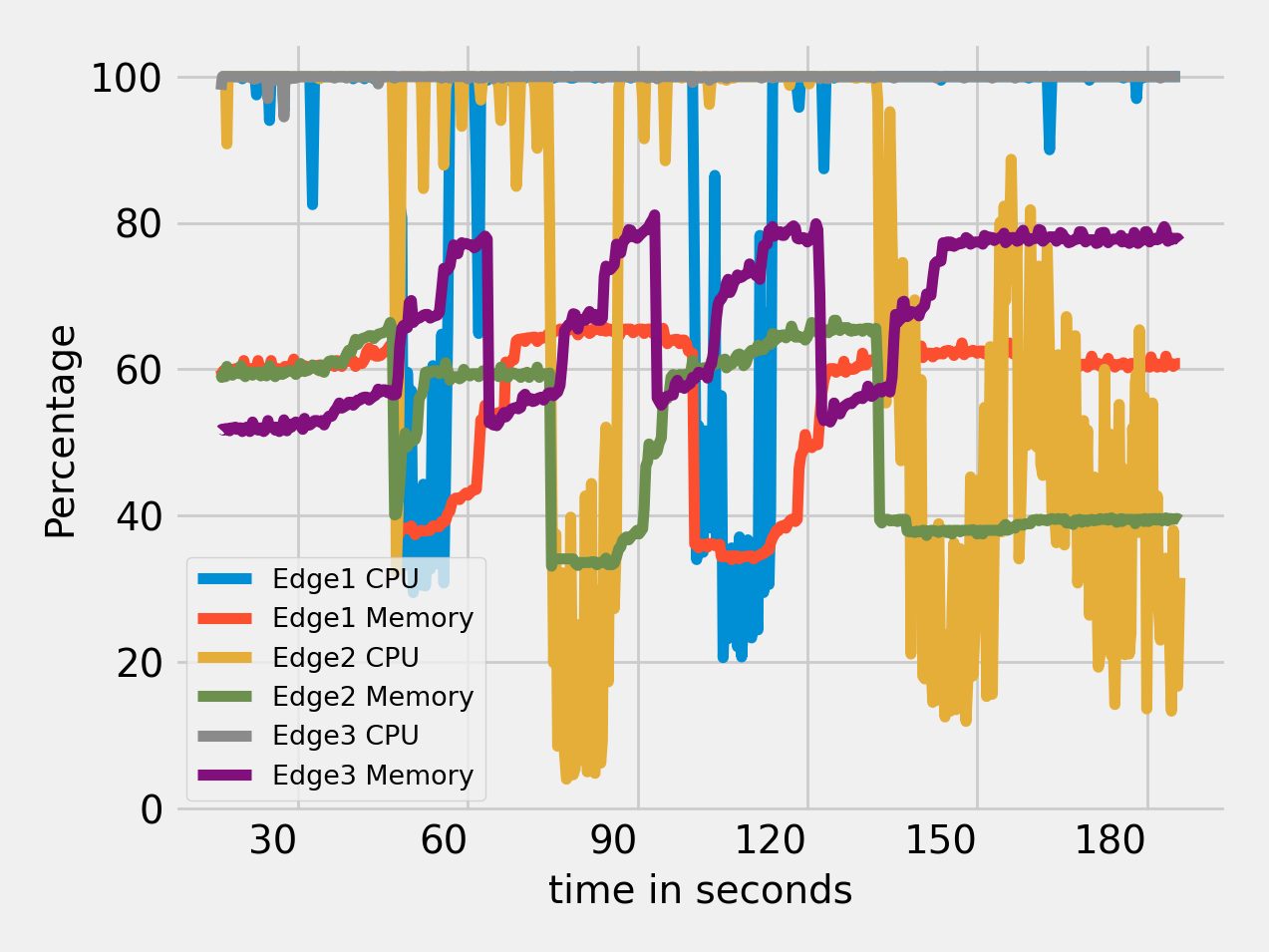}
    \caption{Dynamic Offloading (All-variant)} \label{ts_dy_all_sc2}
\end{subfigure}
\caption{Scenario 2: Resource changes of CPU and memory across devices in a sample trial.}
\label{fig:timeseries_sc2}
\vspace{-2mm}
\end{figure*}

\section{Performance Evaluation} 
We tested the offloading strategies in the aforementioned scenarios. The results are averaged over five independent runs per scenario per strategy. 

\subsection{Effect on Computational Resources}
The experiment results in Table~\ref{table:performance_sc1} (Figs.~\ref{fig:timeseries_sc1} and \ref{fig:timeseries_sc2}) show that our proposed utility-aware dynamic offloading strategy performs optimum resource consumption compared to traditional local and static offloading techniques in both scenarios. Figs.~\ref{ts_local_sc1}, ~\ref{ts_static_sc1}, ~\ref{ts_local_sc2},  ~\ref{ts_static_sc2} show resource contention across the traditional techniques whereas CPU is optimally utilized and kept accessible for other tasks across all the variants of dynamic offloading as shown in Figs. as shown in Fig.~\ref{ts_dy_cpu_sc1}. For scenario 2, since edges are more resource-constrained due to static pre-initialization of mapping and map merging tasks, we see maximum CPU utilization for one of the edges while the rest of them had sufficient processing capability to accommodate more tasks in the sequence. Similarly, The use of memory resources also varies with an obvious curve across dynamic offloading variants, while in static, it remains constant in all edge devices in scenario 1. The memory usage range always fluctuates between 40-80\%, as seen in Fig.~\ref{fig:ts_dy_mem_sc1}. The proposed scheduler distributes the processing load among all the available devices, ensuring tasks are always initialized on the right computing entity based on the changing system conditions as in Fig.~\ref{ts_dy_all_sc1}.

We observe Edge 3 getting assigned most tasks, resulting in higher computational and memory utilization in scenario 1 due to its availability. However, in scenario 2, as the edges are already resource-constrained, the dynamic offloading scheduler ensures computational and memory balance among all variants even under heavy load as shown in Figs.~\ref{ts_dy_cpu_sc2}, \ref{ts_dy_mem_sc2},~\ref{ts_dy_net_sc2} and ~\ref{ts_dy_all_sc2}. This is a clear indication of the efficiency of the scheduler in managing tasks with limited resources. It also proves our claim that resource optimization is achievable with the proposed technique. Additionally, we notice varying throughput on edge devices, with some reaching maximum capacity, implying unequal computational load and reduced throughput due to frequent task hopping and delayed initialization. 

\subsection{Effect on Task Performance}
As shown in Table~\ref{table:performance_sc1}, the map accuracy improves for the robot exploring the environment through the given task through all utility-aware task offloading scheduler variants. We do observe a slight anomaly in expected behavior in scenario 2 as referred in Table~\ref{table:performance_sc1}, which is due to robots collaboratively finishing the task early due to higher throughput, leading to task completion (collection and localization of objects) before fully building the map of the environment. This slight anomaly, however, does not affect the overall performance of the robot in terms of map accuracy. The robots can still build a map with better accuracy, perform navigation, and finish the task of collection and localization of objects, albeit at a faster rate than expected.  

We observe varying task latency across different variants of the proposed scheme, but the "All" variant consistently reduces task latency in both evaluated scenarios. The latency in that variant is reduced by 50\% compared to the local computational approach and 30.2\% compared to the static offloading scheme in scenario 1, whereas it is reduced by 52.6\% and 32\% compared to local and static offloading techniques, respectively, in scenario 2. Since the utility-aware offloading strategy uses multiple distant resources as a pool to execute various tasks in tandem simultaneously, this contributes to shorter job execution times and improved system performance.

Although the proposed scheme shows a drop in the frame rate across all variants of dynamic offloading, this can be attributed to the fact that, at some point, multiple instances of Yolov5 could be running on an edge with more computational resources. Also, it can be attributed to the execution latency related to task initialization of the proposed offloading mechanism, as the execution of a task needs to go through multiple intermediate steps before it reaches its full initialization.  Additionally, the frame rate can be further improved by optimizing the parameters of Yolov5 for each dynamic offloading variant.


\section{Conclusion}
This work presents and evaluates a unique task offloading approach for latency-sensitive edge-based multi-robot systems. The approach proposes a utility-maximization-based task offloading mechanism to minimize total service time and maximize resource consumption by profiling devices (such as CPU utilization, network conditions, and memory needs), as well as performing dynamic computational offloading decision mechanisms. We compare our proposed approach with local task execution and static offloading strategy to analyze task performances by sequencing tasks like mapping, online navigation, object detection and mapping, and map merging. 
Our current and future work focuses on further examining the offloading aspect of the proposed algorithm on different robotic tasks and devising and expanding on the proposed handoff mechanism to realize more efficiency in the system, providing safety guarantees for the robot.

\bibliography{references}

\begin{thebibliography}{10}
\providecommand{\url}[1]{#1}
\csname url@samestyle\endcsname
\providecommand{\newblock}{\relax}
\providecommand{\bibinfo}[2]{#2}
\providecommand{\BIBentrySTDinterwordspacing}{\spaceskip=0pt\relax}
\providecommand{\BIBentryALTinterwordstretchfactor}{4}
\providecommand{\BIBentryALTinterwordspacing}{\spaceskip=\fontdimen2\font plus
\BIBentryALTinterwordstretchfactor\fontdimen3\font minus \fontdimen4\font\relax}
\providecommand{\BIBforeignlanguage}[2]{{%
\expandafter\ifx\csname l@#1\endcsname\relax
\typeout{** WARNING: IEEEtran.bst: No hyphenation pattern has been}%
\typeout{** loaded for the language `#1'. Using the pattern for}%
\typeout{** the default language instead.}%
\else
\language=\csname l@#1\endcsname
\fi
#2}}
\providecommand{\BIBdecl}{\relax}
\BIBdecl

\bibitem{tahir2022analog}
N.~Tahir and R.~Parasuraman, ``Analog twin framework for human and ai supervisory control and teleoperation of robots,'' \emph{IEEE Transactions on Systems, Man, and Cybernetics: Systems}, vol.~53, no.~5, pp. 2616--2628, 2023.

\bibitem{spatharakis2022resource}
D.~Spatharakis, M.~Avgeris, N.~Athanasopoulos, D.~Dechouniotis, and S.~Papavassiliou, ``Resource-aware estimation and control for edge robotics: A set-based approach,'' \emph{IEEE Internet of Things Journal}, 2022.

\bibitem{uysal2021semantic}
E.~Uysal, O.~Kaya, A.~Ephremides, J.~Gross, M.~Codreanu, P.~Popovski, M.~Assaad, G.~Liva, A.~Munari, T.~Soleymani \emph{et~al.}, ``Semantic communications in networked systems,'' \emph{arXiv preprint arXiv:2103.05391}, 2021.

\bibitem{9289887}
K.~Antevski, M.~Groshev, G.~Baldoni, and C.~J. Bernardos, ``Dlt federation for edge robotics,'' in \emph{2020 IEEE Conference on Network Function Virtualization and Software Defined Networks (NFV-SDN)}, 2020, pp. 71--76.

\bibitem{antevski2018enhancing}
K.~Antevski, M.~Groshev, L.~Cominardi, C.~J. Bernardos, A.~Mourad, and R.~Gazda, ``Enhancing edge robotics through the use of context information,'' in \emph{Proceedings of the Workshop on Experimentation and Measurements in 5G}, 2018, pp. 7--12.

\bibitem{9677930}
Y.~Wang, W.~Wang, D.~Liu, X.~Jin, J.~Jiang, and K.~Chen, ``Enabling edge-cloud video analytics for robotics applications,'' \emph{IEEE Transactions on Cloud Computing}, pp. 1--1, 2022.

\bibitem{parasuraman2017new}
R.~Parasuraman, S.~Caccamo, F.~B{\aa}berg, P.~Ogren, and M.~Neerincx, ``A new ugv teleoperation interface for improved awareness of network connectivity and physical surroundings,'' \emph{Journal of Human-Robot Interaction}, vol.~6, no.~3, pp. 48--70, 2017.

\bibitem{8010408}
Y.~Sahni, J.~Cao, S.~Zhang, and L.~Yang, ``Edge mesh: A new paradigm to enable distributed intelligence in internet of things,'' \emph{IEEE Access}, vol.~5, pp. 16\,441--16\,458, 2017.

\bibitem{10.1145/3004010.3004032}
S.~Dey and A.~Mukherjee, ``Robotic slam: A review from fog computing and mobile edge computing perspective,'' in \emph{Adjunct Proceedings of the 13th International Conference on Mobile and Ubiquitous Systems: Computing Networking and Services}, ser. MOBIQUITOUS 2016.\hskip 1em plus 0.5em minus 0.4em\relax New York, NY, USA: Association for Computing Machinery, 2016, p. 153–158.

\bibitem{9279239}
W.-Z. Zhang, I.~A. Elgendy, M.~Hammad, A.~M. Iliyasu, X.~Du, M.~Guizani, and A.~A.~A. El-Latif, ``Secure and optimized load balancing for multitier iot and edge-cloud computing systems,'' \emph{IEEE Internet of Things Journal}, vol.~8, no.~10, pp. 8119--8132, 2021.

\bibitem{dechouniotis2022edge}
D.~Dechouniotis, D.~Spatharakis, and S.~Papavassiliou, ``Edge robotics experimentation over next generation iiot testbeds,'' in \emph{NOMS 2022-2022 IEEE/IFIP Network Operations and Management Symposium}.\hskip 1em plus 0.5em minus 0.4em\relax IEEE, 2022, pp. 1--3.

\bibitem{huang2021efficient}
X.~Huang, R.~Yu, D.~Ye, L.~Shu, and S.~Xie, ``Efficient workload allocation and user-centric utility maximization for task scheduling in collaborative vehicular edge computing,'' \emph{IEEE Transactions on Vehicular Technology}, vol.~70, no.~4, pp. 3773--3787, 2021.

\bibitem{tahir2023percom}
N.~Tahir and R.~Parasuraman, ``Mobile robot control and autonomy through collaborative twin,'' in \emph{2023 IEEE International Conference on Pervasive Computing and Communications Workshops and other Affiliated Events (PerCom Workshops)}, 2023, pp. 558--563.

\bibitem{mushunuri2017resource}
V.~Mushunuri, A.~Kattepur, H.~K. Rath, and A.~Simha, ``Resource optimization in fog enabled iot deployments,'' in \emph{2017 Second International Conference on Fog and Mobile Edge Computing (FMEC)}.\hskip 1em plus 0.5em minus 0.4em\relax IEEE, 2017, pp. 6--13.

\bibitem{4804363}
A.~Birk, S.~Schwertfeger, and K.~Pathak, ``A networking framework for teleoperation in safety, security, and rescue robotics,'' \emph{IEEE Wireless Communications}, vol.~16, no.~1, pp. 6--13, 2009.

\bibitem{huang2022edge}
P.~Huang, L.~Zeng, X.~Chen, K.~Luo, Z.~Zhou, and S.~Yu, ``Edge robotics: Edge-computing-accelerated multi-robot simultaneous localization and mapping,'' \emph{IEEE Internet of Things Journal}, 2022.

\bibitem{ben2022edge}
A.~J. Ben~Ali, M.~Kouroshli, S.~Semenova, Z.~S. Hashemifar, S.~Y. Ko, and K.~Dantu, ``Edge-slam: Edge-assisted visual simultaneous localization and mapping,'' \emph{ACM Transactions on Embedded Computing Systems}, vol.~22, no.~1, pp. 1--31, 2022.

\bibitem{chinchali2021network}
S.~Chinchali, A.~Sharma, J.~Harrison, A.~Elhafsi, D.~Kang, E.~Pergament, E.~Cidon, S.~Katti, and M.~Pavone, ``Network offloading policies for cloud robotics: a learning-based approach,'' \emph{Autonomous Robots}, vol.~45, no.~7, pp. 997--1012, 2021.

\bibitem{xu2020edge}
J.~Xu, H.~Cao, D.~Li, K.~Huang, C.~Qian, L.~Shangguan, and Z.~Yang, ``Edge assisted mobile semantic visual slam,'' in \emph{IEEE INFOCOM 2020-IEEE Conference on Computer Communications}.\hskip 1em plus 0.5em minus 0.4em\relax IEEE, 2020, pp. 1828--1837.

\bibitem{zhang2019resource}
X.~Zhang, Y.~Zhong, P.~Liu, F.~Zhou, and Y.~Wang, ``Resource allocation for a uav-enabled mobile-edge computing system: Computation efficiency maximization,'' \emph{IEEE Access}, vol.~7, pp. 113\,345--113\,354, 2019.

\bibitem{8314091}
H.~Liu, S.~Liu, and K.~Zheng, ``A reinforcement learning-based resource allocation scheme for cloud robotics,'' \emph{IEEE Access}, vol.~6, pp. 17\,215--17\,222, 2018.

\bibitem{penmetcha2021deep}
M.~Penmetcha and B.-C. Min, ``A deep reinforcement learning-based dynamic computational offloading method for cloud robotics,'' \emph{IEEE Access}, vol.~9, pp. 60\,265--60\,279, 2021.

\bibitem{penmetcha2021predictive}
M.~Penmetcha, S.~S. Kannan, and B.-C. Min, ``A predictive application offloading algorithm using small datasets for cloud robotics,'' in \emph{2021 IEEE International Conference on Systems, Man, and Cybernetics (SMC)}.\hskip 1em plus 0.5em minus 0.4em\relax IEEE, 2021, pp. 132--139.

\bibitem{9511225}
M.~Afrin, J.~Jin, A.~Rahman, A.~Gasparri, Y.-C. Tian, and A.~Kulkarni, ``Robotic edge resource allocation for agricultural cyber-physical system,'' \emph{IEEE Transactions on Network Science and Engineering}, pp. 1--1, 2021.

\end{thebibliography}
\bibliographystyle{IEEEtran}
\end{document}